\documentclass[num-refs]{wiley-article}

\makeatletter
\newif\ifanonymous
\if@blindreview\anonymoustrue\else\anonymousfalse\fi
\makeatother

\usepackage{siunitx}
\usepackage{todonotes}
\usepackage{listings}
\usepackage{dirtytalk}
\usepackage{multicol}
\usepackage{wasysym}

\newcommand{\student}[1]{\textsc{#1}}

\newcommand{\yesSym}{\CIRCLE}
\newcommand{\noSym}{\Circle}
\newcommand{\partSym}{\LEFTcircle}

\papertype{Original Article}
\paperfield{Journal Section}

\title{You're (Not) My Type -- Can LLMs Generate Feedback of Specific Types for Introductory Programming Tasks?}

\author[1\authfn{1}]{Dominic Lohr}
\author[2]{Hieke Keuning}
\author[3]{Natalie Kiesler}

\affil[1]{Friedrich-Alexander-Universität Erlangen-Nürnberg}
\affil[2]{Utrecht University}
\affil[3]{Nuremberg Tech}

\corraddress{Dominic Lohr, Professorship for Computer Science Education, Friedrich-Alexander-Universität Erlangen-Nürnberg, 91058, Germany}
\corremail{dominic.lohr@fau.de}

\presentadd[\authfn{1}]{Department of Computer Science, Martensst. 3, Friedrich-Alexander-Universität Erlangen-Nürnberg, Germany}

\runningauthor{Lohr et al.}

\newcommand{\lb}[1]{#1} 
\newcommand{\lbb}[1]{#1} 

\newcommand{\gpt}[1]{\textit{``#1''}}
\newcommand{\snpt}[1]{\texttt{#1}}

\lstdefinestyle{mystyle}{
    basicstyle=\ttfamily\footnotesize,
    breakatwhitespace=false,         
    breaklines=true,                 
    captionpos=t,                    
    keepspaces=true,                 
    numbers=left,                    
    numbersep=5pt,                  
    tabsize=2,
    frame=single
}
\begin{document}

\begin{frontmatter}
\maketitle

\begin{abstract}
\noindent \textbf{Background}: Feedback as one of the most influential factors for learning has been subject to a great body of research. It plays a key role in the development of educational technology systems and is traditionally rooted in deterministic feedback defined by experts and their experience. However, with the rise of generative AI and especially Large Language Models (LLMs), we expect feedback as part of learning systems to transform, especially for the context of programming. In the past, it was challenging to automate feedback for learners of programming. LLMs may create new possibilities to provide richer, and more individual feedback than ever before.

\noindent \textbf{Objectives}: This paper aims to generate specific types of feedback for introductory programming tasks using LLMs. We revisit existing feedback taxonomies to capture the spe\-cifics of the generated feedback, such as randomness, uncertainty, and degrees of variation. 

\noindent \textbf{Methods}: We iteratively designed prompts for the generation of specific feedback types (as part of existing feedback taxonomies) in response to authentic student programs. We then evaluated the generated output and determined to what extent it reflected certain feedback types. 

\noindent \textbf{Results and Conclusion}: The present work provides a better understanding of different feedback dimensions and characteristics. The results have implications for future feedback research with regard to, for example, feedback effects and learners' informational needs. It further provides a basis for the development of new tools and learning systems for novice programmers including feedback generated by AI.

\keywords{feedback, introductory programming, generative AI, GenAI, ChatGPT, GPT-4, large language models, feedback classification, feedback types}

\end{abstract}
\end{frontmatter}

\vspace*{-0.9cm}
\section{Introduction}

Learning to program involves much more than just understanding the syntax and semantics of a programming language. It is therefore not surprising that it has been characterized as a challenging process~\cite{qian2017students, medeiros2018systematic,luxton2018introductory,kiesler2024modeling}. Even experienced programmers regularly encounter errors in their code -- a testament to the inherent complexity of programming. Due to experience and years of practice, programming experts have an arsenal of debugging techniques at their hands. Novice programmers, however, lack this experience, causing them to be more dependent on external support, e.g., from educators, peers, or tools. This is particularly true at the very beginning of the learning process.

Feedback -- as a means of external support -- is one of the most influencing factors of learning success~\cite{hattieVisibleLearningSynthesis2009}. In the context of programming education, elaborated feedback has not yet been implemented at scale, for example, as a feature of learning environments~\cite{jeuring2022wgfullreport}. Common systems (e.g., Codewars, Codecademy, Python Tutor, W3Schools) often tend to provide simple feedback indicating the correctness of the code or pointing out errors without addressing the underlying causes or how to fix these errors~\cite{jeuring2022wgfullreport}. In contrast, adaptive feedback addressing the causes of errors and not just their manifestation in the code is considered more valuable by learners~\cite{lohrCriteriaValuableAutomatic2021}.

The recent emergence of advanced Natural Language Processing (NLP) techniques, generative AI (GenAI), and particularly the broad availability of Large Language Models (LLMs), offers new opportunities for improving and elaborating feedback in programming education. Prior research has started to investigate the potential of LLMs to generate, for example, code explanations, enhanced error messages, and thus formative programming feedback \cite{azaiz2024feedback,leinonen2023using,Sarsa2022,macneil2022experiences,azaiz2023aienhanced,Bengtsson_Kaliff_2023,kiesler2023exploring}. However, these studies are usually based on simple help-requests as phrased by students. Moreover, the rich diversity of feedback types with regard to its content has not yet been addressed.

In this paper, we explore to what extent the feedback generated by LLMs can be controlled and how its quality can be assessed. Our experiments pursue the following goals: (1) design prompts to generate specific types of feedback from a specific feedback taxonomy for the context of programming (see~\cite{keuning2018}), and (2) evaluate the adequacy and quality of the generated feedback types. To address these goals, we used an iterative process to develop a prompt to generate different types of feedback. We used this prompt and a dataset of authentic student solutions (correct ones and with errors) as input, aiming to generate different types of feedback for these student solutions. The same dataset has been used in related work~\cite{kiesler2023exploring} to generate feedback, but this related study did not apply ``prompt engineering techniques'' -- meaning the input was simple and the feedback generated by the LLM was characterized with regard to its content, quality, and other criteria.

Our paper is structured as follows. In Section \ref{sec:bg}, we describe related work on the role of feedback in learning in general, and how feedback has been automated in the context of computing education. We also outline recent research on the emergence of generative AI in computing education and summarize studies employing GenAI for the generation of feedback in programming education. 
We describe the method of our work in Section \ref{sec:method}, which includes the selection of datasets with (erroneous) student programs, the prompt design process, and the approach for characterizing the feedback output. We present and discuss the results in Section \ref{sec:res}, and the limitations in Section \ref{sec:lim}. In Section \ref{sec:implications}, we explore the implications for educators, researchers, students, and tool designers, which is followed by conclusions in Section \ref{sec:concl}.

\section{Background and related work}
\label{sec:bg}

To contextualize the present work, we introduce the role of feedback as part of students' learning process before focusing on feedback types for programming tasks and the recent role of generative AI in programming education.

\subsection{The role of feedback in the learning process and the programming domain}
\label{sec:automatedfb}

Feedback has been the subject of extensive research in a great number of disciplines. According to a meta-analysis, it is one of the most influential factors for learning~\cite{hattieVisibleLearningSynthesis2009}. At the same time, it is crucial to consider how feedback is delivered and presented~\cite{narciss2008feedback}. In general, the goal of giving feedback is to reduce the gap between the student's current performance and their desired goal~\cite{hattietimperley2007}, and to modify their thinking and behavior \cite{shute}. Research shows the importance of giving detailed feedback, as opposed to simple messages only indicating whether a solution is correct or incorrect. Feedback may also be classified as \emph{summative} (focusing on performance after submission to educators) or \emph{formative} (focusing on the learning process during the execution of a task or a course). In the present work, we refer to formative feedback provided to students trying to solve a task or assignment as part of their coursework.

According to \citeauthor{hattietimperley2007} \cite{hattietimperley2007}, effective feedback should answer the questions `where am I going' (feed up), `how am I going' (feed back), and `where to next' (feed forward). Each of these questions can be applied to four levels: (1) task, (2) process, (3) self-regulation, and (4) self. 
\citeauthor{shute} \cite{shute} recommends formative feedback to be non-evaluative, supportive, timely, and specific. 

The literature distinguishes various dimensions for the design of feedback~\cite{narciss2008feedback, shute}.
Narciss \cite{narciss2008feedback} outlines \textit{function} (cognitive, meta-cognitive, and motivational), \textit{presentation} (timing, number of tries, adaptability, modality), and \textit{content} of feedback. Each of these dimensions can influence the feedback's effects. Regarding content, Narciss classified feedback components for digital learning environments based on which aspects of the instructional context the feedback addresses, such as task constraints, mistakes, and procedural knowledge.

Narciss \cite{narciss2006} summarizes feedback into \textit{simple} and \textit{elaborated} types. The latter provides more detailed information to the learner to analyze and improve their work. \mbox{\citeauthor{keuning2018} \cite{keuning2018}} have applied and extended this categorization for the programming domain. Table 1 provides an overview of the different feedback types with a description of how they are applied to the context of programming. The feedback types we used to analyze the data are marked with an asterisk (${}^*$).

\begin{table}[!htbp]
\caption{Feedback Types \cite{keuning2018}.}
\label{tab:types}
\begin{center}
    
\begin{tabular}{p{1.8cm} p{2.2cm} p{9cm}}
\hline
\textbf{Label} & \textbf{Name} & \textbf{Description} \\
\hline\hline

\textit{simple types} && \\ \hline\hline
KR${}^*$ & Knowledge of\newline result & a simple indication of `correct' or `incorrect'. \citeauthor{keuning2018} define `correct' in the context of a programming task as one or more of the following elements: 1) it passes all tests, 2) it is equal to a model program, and/or 3) it satisfies one or more constraints. \\
KCR & Knowledge of\newline correct result & A description or an indication of the correct response (e.g. providing the learner a model solution)  \\
KP${}^*$ & Knowledge of\newline performance & Summative feedback on the achieved performance level regarding a set of tasks (e.g. ‘6 out of 10 tasks correct’ or `75\% correct'.)  \\ \hline \hline
\textit{elaborate types} &&\\ \hline \hline
KTC${}^*$ & Knowledge about\newline task constraints & Focuses on the task itself. This feedback could be about certain constraints (not using library functions, or enforcing a specific approach, such as recursion), or more general hints on how to approach the task without taking the student's current program into account.  \\
KC${}^*$ & Knowledge about\newline concepts & Contains explanations on topics and concepts relevant to the exercise, or examples illustrating these concepts.  \\
KM${}^*$ & Knowledge about\newline mistakes & Points out mistakes, by means of showing syntactic or semantic errors detected by a compiler (not exercise-specific), test cases that fail, or solution errors (logical or runtime). Other subcategories focus on style and performance-related issues. This content can be \textit{basic}, containing a number (e.g. showing the number of failed tests), location or short identifier; or with a \textit{detailed} description.\\
KH${}^*$ & Knowledge on\newline how to proceed & Hints on how to correct errors, next-step hints, and hints on how to improve the style of a semantically correct program. \\
KMC & Knowledge about\newline meta-cognition & Feedback related to problem-solving strategies and the student's awareness of their own performance. \\

\hline
\multicolumn{3}{c}{${}^*$ used for data analysis}
\end{tabular}
\end{center}

\end{table}

\citeauthor{keuning2018} \cite{keuning2018} labeled the feedback of more than 100 available programming learning tools using these categories, showing that feedback provided by these tools (based on papers up until 2015) mostly focuses on pointing out mistakes. 
The authors omitted the simple feedback types in their study. However, the simple types of feedback have been used in many learning tools, for example, to indicate correctness or to show the fully correct solution. At the same time, other research has shown that binary feedback may not be beneficial to learning \cite{kyrilov2016students}.

The presented classification has been used in other studies (e.g.~\cite{jeuring2022wgfullreport, brocker2023investigating,lohrLetThemTry2024,kiesler2023investigating}). \citeauthor{jeuring2022wgfullreport}~\cite{jeuring2022wgfullreport}, for example, investigated the feedback types implemented in well-known online programming learning environments. They found that these environments mostly provide simple feedback and compiler or test-based feedback. Moreover, they found that educators as experts would provide different types of feedback, i.e., more elaborate feedback types (see also ~\cite{lohrLetThemTry2024}). 

The described feedback types by Keuning et al.~\cite{keuning2018} are closely related to the types of issues, errors, and misconceptions students may have or produce. \citeauthor{qian2017students}'s review of misconceptions in introductory programming~\cite{qian2017students} discusses issues that arise in three main categories related to programming knowledge (based on \citeauthor{bayman1988using}~\cite{bayman1988using}), referring to syntactic, conceptual, or strategic knowledge. Syntactic errors are often caught by the compiler. However, it is well known that students struggle with compiler error messages \cite{Becker2019Unhelpful}. Conceptual problems can be caused by misconceptions regarding mental models of code execution and computer systems (e.g., ~\cite{kiesler2022mental}). They are much harder to address but may be mitigated by certain feedback types (KC, KM subtypes on solution errors and test failures, and KH error correction feedback). Strategic problems concern planning, writing, and debugging programs as a solution to a novel problem. They require both syntactic and conceptual knowledge as a prerequisite.

In addition, non-functional aspects of programming have been gaining more attention in courses and assessments, such as code quality, style, and performance. 
In other cases, a program might be error-free but incomplete, and the student needs help with the next step. Hence, students' informational needs can greatly vary.

\subsection{Using generative AI for feedback in programming education}

In the past decades, many different techniques have been applied to automatically generate feedback with the goal of supporting educators and learners at scale (e.g., in Massive Open Online Courses)~\cite{messer2024automated,paiva2022automated,keuning2018}. While automated testing is still one of the most common ways to generate feedback, several data-driven methods have emerged. However, with LLMs and GenAI tools becoming available to the public at the end of 2022, the possibility of generating a diverse range of feedback has become a reality and a true option for both educators and learners. 

Accordingly, GenAI and LLMs have been extensively investigated within the last two years, especially in the context of programming education~\cite{prather2023wgfullreport,prather2024wgfull,becker2023generative}. Early papers focused on analyzing how well LLMs could solve well-known (introductory) programming problems~\cite{Finnie2022,denny2023conversing,wermelinger2023using,cipriano2023gpt-3,kiesler2023large}. Recently, researchers have shifted their focus towards investigating the use of LLMs to enhance teaching and learning how to program, and their integration into the classroom~\cite{scholl2024analyzing,scholl2024hownovice,prather2024wgabstract}. OpenAI's GPT-4 model's capabilities in identifying errors within programming code have been explored~\cite{wu2023large}, as well as its performance in (beginner and intermediate) \texttt{Python} course exams~\cite{savelka2023large}, answering various types of questions~\cite{joshi2024solvingquestions}, or for automatic program repair and refactoring~\cite{ishizue2024programrepair}.

Another potential application of LLMs is the generation of code explanations. \citeauthor{macneil2022experiences}~\cite{macneil2022experiences} showed that students perceive generated line-by-line code explanations as helpful. In another study~\cite{leinonen2023comparing}, students rated the code explanations of the LLM better on average than those created by students. Especially the level of understanding and accuracy were rated higher, and students did not show negative affection towards the generated, non-human feedback. 

Due to GenAI tools and LLMs, it is thus possible to generate feedback for (novice) learners of programming. Several studies have investigated the feedback generated by LLMs. 
For example, \citeauthor{balse2023investigating}~\cite{balse2023investigating} used final exam submissions to seven exercises as input to GPT-3, and evaluated the response on correctness and type of feedback. They found that 71\% of the feedback was accurately checked as correct, and around 62\% contained correct critiques and suggestions. The authors identified four themes in the feedback that mostly target the constructs on which feedback was given: `generic feedback', `variable initialization and updates', `conditionals', and `iteration'.

\citeauthor{pankiewicz2023large}~\cite{pankiewicz2023large} implemented GPT-3.5 feedback inside an existing automated assessment tool, targeted at \texttt{C\#} submissions to 38 exercises. The students rated the feedback on a Likert scale, and their performance, time-on-task, and self-reported affective state were measured and compared to a control group that did not receive the hints. Almost half of the students were positive about the feedback, which was in Polish, but some students were less happy with the hints. The authors also found that the experimental group with GPT hints solved tasks faster and submitted correct solutions more often. When they had to solve problems without hints afterwards, they initially struggled, but caught up with the control group in the end. The prompt contained the Polish exercise description, the student's code, information about failed tests and compiler errors, and based on the latter, a prompt with instructions. An example of this prompt was given in the paper, which instructed not to give away the solution and only give hints, and to highlight certain elements (e.g. keywords, variable names, line numbers) using html-tags.  

Related work by \citeauthor{kiesler2023exploring}~\cite{kiesler2023exploring} generated feedback for various \texttt{Java} programming exercises using ChatGPT 3.5, and a very simple prompt (which students might use). The prompt consisted of the question \say{What is wrong with my code?} followed by the student submission to an introductory programming task. The generated output was qualitatively analyzed with regard to its content, quality, and other elements (see Table \ref{tab:legend}). The authors found that several LLM-generated feedback messages contained misleading information~\cite{kiesler2023exploring}. Some of these were due to the fact that the prompt did not contain any information about the task or special task constraints. For example, it was not permitted to use \texttt{Java} libraries, although this was occasionally requested or included in the LLM's feedback.

\begin{table}[htb]
\caption{Characteristics of LLM feedback messages \cite{kiesler2023exploring}.}
\label{tab:legend}
\begin{center}
    
\begin{tabular}{l p{8cm} l}
\hline
\textbf{Label} & \textbf{Meaning} & \textbf{Value options} \\
\hline\hline

\textit{Content} && \\ \hline\hline
INFO & Requesting more information & Y/N \\
STYLE & Stylistic suggestion & Y/N  \\
CAUSE & Textual explanation cause of error & Y/N  \\
FIX & Textual explanation fix of error & Y/N  \\
CODE & Code provided. A snippet could be a (few) statement(s), or an expression, but not a single variable name.  & Y/Snippet/N  \\ 
EXA & Illustrating examples & Y/N\\\hline\hline
\textit{Quality} &&\\ \hline\hline
COMP & Code, if provided, compiles (or in case of a snippet, is correct) & Y/N/n.a. \\
MIS & Contains misleading information & Y/N \\
UNC & Expression of uncertainty & Y/N \\
PERS & It is personalized to the student’s code, such as using their variable names or referring to their algorithm & Y/N \\
COMPL & It matches the task description & Y/N \\

\textit{Other} &&\\ \hline
META & Meta-cognitive elements & Y/N \\ 
MOT & Motivational elements & Y/N\\

\hline
\end{tabular}
\end{center}

\end{table}

\citeauthor{roest2023nextstep} \cite{roest2023nextstep} focused on generating next-step hints for introductory \texttt{Python} exercises. They iteratively developed a prompt for GPT 3.5 to generate a short hint on what to do next (i.e. when a student is working towards a solution). The prompt contained the task description, and specific keywords in a relatively short instruction. This prompt was then used in an online programming tool to deliver hints on the students' requests. A small experiment with students was conducted, as well as an evaluation of generated hints with experts, using the following 9 criteria: \emph{type}, \emph{level of detail}, \emph{information}, \emph{personalized}, \emph{appropriate}, \emph{specific}, \emph{misleading information}, \emph{tone}, and \emph{length}. 
Overall, the feedback from students and experts regarding the hints was positive. However, there were instances of misleading information, and the hints were found to lack detail, particularly as students neared the completion of the exercises.
\citeauthor{xiao2024exploring} \cite{xiao2024exploring} also explored hint generation. Specifically, they measured the effects of four levels of hints, from a generic text hint to concrete code solutions.

\citeauthor{azaiz2023aienhanced} \cite{azaiz2023aienhanced} characterized the feedback generated by GPT3 (text-davinci-003) for both correct and incorrect submission to two \texttt{Java} programming exercises. They used a simple prompt along with the task description and the student code. They studied the responses' length (the difference between the two exercises was statistically significant), whether it correctly identified correct solutions (in 73\% of the cases, comparable to \citeauthor{balse2023investigating} \cite{balse2023investigating} who employed the same model), and the characteristics of its content (code elements, personalization, style suggestions, correctness of suggestions, specification compliance, and fault localization). Submissions were finally clustered into three categories: syntactically and functionally correct (SCFC), syntactically correct but functionally not correct (SCFI, not meeting task requirements), and syntactically and functionally incorrect (SIFI). Several issues with the LLM-generated feedback are reported, such as giving suggestions that do not align with the requirements. The authors~\cite{azaiz2023aienhanced} conclude that \say{currently, it is not good enough for summative grading}. However, they see the potential of LLMs to support teaching assistants in grading.

In a recent follow-up study by Azaiz et al.~\cite{azaiz2024feedback}, GPT-4 Turbo's feedback in response to the dataset with the same 55 student programming submissions as in prior work~\cite{azaiz2023aienhanced} were evaluated. They qualitatively analyzed the output with regard to its feedback content and structure, code representation, corrections, and correction types, suggested optimizations and coding style, as well as inconsistencies and redundancies. They conclude that all of the feedback generated by GPT-4 is personalized and significantly better in terms of its accuracy compared to previous GPT versions~\cite{azaiz2024feedback}. Moreover, they note that following all of the LLM's suggestions and corrections would have resulted in fully correct solutions (except for two cases). At the same time, only 52\% of the feedback and all of its details were fully correct and complete. 

\citeauthor{nguyen2024usinggpt4feedback}~\cite{nguyen2024usinggpt4feedback} present their experience with using GPT-4 to provide tiered, formative feedback to \texttt{Java} code and pseudocode in the context of a course on algorithms and data structures. Their goal was to elicit feedback on conceptual understanding, syntax, and time complexity, as well as next-step hints or guiding questions. They analyzed the feedback regarding its accuracy and usefulness and its correctness, concluding that GPT-4 generally provided accurate feedback, which was perceived as useful in guiding students’ next steps. However, they also noted a varying performance across the output for the 113 submissions to four programming problems.

\citeauthor{phung2024automating} \cite{phung2024automating} have developed a new technique to provide feedback on buggy programs, incorporating symbolic data such as failed tests and a fixed program to generate LLM-based feedback, with a validation step to ensure higher quality.
\citeauthor{koutcheme2024open} \cite{koutcheme2024open} have explored the capabilities of open-source LLMs to generate feedback, and used GPT to automatically assess the feedback quality.

Finally, several studies have focused on using LLMs to enhance compiler error messages. \citeauthor{wang2024large} \cite{wang2024large} used GPT to enhance \texttt{Python} error messages. They conducted a large-scale randomized control trial in which they compared six different approaches, of which two approaches used GPT-3. They found that the GPT-enhanced messages were most effective; students resolved the error in fewer attempts, and they repeated the error less often. \citeauthor{taylor2024dcc} \cite{taylor2024dcc} present a tool that integrates an LLM into a \texttt{C} compiler, providing student-friendly error messages that help them solve the problem. They deployed the tool at a large scale and analyzed the quality of the generated messages. \citeauthor{kimmel2024enhancing} \cite{kimmel2024enhancing} used GPT-4 to give feedback on compiler, runtime, and logic errors, which they deploy in an automated assessment tool. They observed mixed opinions from students and argue that the UI design of such a tool influences how the feedback is perceived.

\section{Method}
\label{sec:method}

\subsection{Research motivation}

Related research had the goal to generate \say{general} feedback in response to a task, student solution, or the question of how to proceed. These studies concluded that LLM-generated feedback messages may contain misleading information, incorrect corrections, and lack of details~\cite{azaiz2024feedback,azaiz2023aienhanced,kiesler2023exploring,roest2023nextstep} in response to student programming submissions. At the same time, LLM-generated feedback is personalized and can address several issues and aspects of an input (e.g., conceptual knowledge, the task itself, mistakes, hints on how to improve, or even meta-cognitive strategies).

To date, learning environments for programming mostly provide a limited set of feedback types, and they hardly give elaborated, let alone personalized feedback~\cite{jeuring2022wgfullreport}. Therefore, it is the goal of the present work to investigate how and to what extent we can generate specific, elaborated feedback types by using a state-of-the-art LLM (e.g., GPT-4). 

In addition, currently available programming feedback classifications like the one by \citeauthor{keuning2018}~\cite{keuning2018} were developed at a time when feedback generation options were limited and complex to automate. In the LLM era, generative AI may provide us with a richer array of possibilities, implying the need to revisit this taxonomy to showcase these possibilities.

\subsection{Scope and research questions}

The present work focuses on giving feedback to students working on a single programming exercise while limiting the scope to the following aspects: (1) The student's program for which we want to provide feedback is (almost) complete. We do not consider partial solutions from the early phases of the problem-solving process. (2) The goal of this work is to critically analyze the input provided to the LLM, and to identify suggestions on how to improve and expand the input in an educational setting. This scenario differs from related work employing LLMs for code explanations, fault localization and repair for a professional setting, and enhancing existing compiler error messages. (3) We focus on formative, elaborated feedback, but also investigate some simple elements such as KR as a component of more extended feedback. The following research questions guide our work:
\begin{enumerate}
    \item [\textbf{RQ 1}] To what extent can feedback of a specific type be generated using LLMs for a given (faulty) program?
    \item [\textbf{RQ 2}] What are the characteristics of the generated feedback?
\end{enumerate}

\subsection{(Re-)Use of Datasets}

\subsubsection{Dataset Used for Prompt Development}
To create a complete and comprehensive prompt, we, first of all, made sure that the prompt we intended to use would include all relevant information as input to the LLM. Therefore, we selected a dataset that encompassed all the information available about its setting and context. For example, the task description, a model solution, and templates (e.g., class skeleton) students may have received. Precisely, we used the dataset from a related study, published at the 2023 Frontiers in Education (FIE) conference~\cite{kiesler2023exploring}. The respective data was gathered via weekly exercises in a (large) introductory programming course in \texttt{Java}. Three exercises were selected from this set (Physics, Triangles, and NegaFib in Table \ref{tab:tasks}).  
For all but the NegaFib exercise a template with a class skeleton containing empty methods was provided to the students. We made some changes to the Triangle task by merging its two subtasks into one task. This was because we wanted to prompt the LLM to generate the complete solution, but also all additional information (e.g., related programming concepts). 
This way, the use case was more similar to students solving the complete task as part of their coursework.

\begin{table}[h!]
\renewcommand*{\arraystretch}{1.4}
\caption{Exercise overview for prompt development and evaluation.}\label{tab:tasks}

\begin{tabular}{p{1cm} p{5cm} p{2cm} p{1.2cm} p{3cm}}
\thead{Name} & \thead{Description} & \thead{Concepts} & \thead{Language} & \thead{used for} \\
\hline
Physics   &  Write methods for the given thermal equations of state using the constants provided.  & class, methods, calculations, constants & Java  & prompt development, \newline evaluation
\\

Triangles & Write methods to classify a triangle (1) by its sides (equilateral, isosceles, or scalene), depending on whether all three sides, exactly two sides, or no sides are equal, and (2) by its angles (acute, right, or obtuse) depending on whether all three angles are less than 90$^{\circ}$, one angle is exactly 90$^{\circ}$, or one angle is greater than 90$^{\circ}$. & class, methods, calculations, conditionals, enums  & Java & prompt development, \newline evaluation
\\

NegaFib & Compute the negative Fibonacci sequence for negative inputs without using  classes or methods from the Java API.     & methods,\newline conditionals,\newline recursion    & Java & prompt development, \newline  evaluation
\\
\hline
Fibonacci & Compute the n-th Fibonacci number using recursion. & methods, \newline conditionals, \newline recursion & Java & evaluation \\
Password & A function asking the user for a password until they enter the correct one (given as parameter). & methods, parameters, loops, conditionals, input/output & Dart & evaluation \\
MaxOf3 & Print the largest of three numbers in the input. & I/O,\newline conditionals & Python \newline Kotlin,\newline Java, C++ & evaluation \\

\hline    
\end{tabular}
\end{table}




As an initial set for designing the prompt, we selected one student submission for each exercise that contained multiple errors, such as syntax errors, semantic errors and logic errors (for an overview see Table \ref{tab:submissions}).

\begin{table}[h!]
\renewcommand*{\arraystretch}{1.4}
\caption{Overview of selected student submissions}\label{tab:submissions}

\begin{tabular}{p{1.3cm} p{1.3cm} p{1.3cm} p{1.3cm} p{6cm}}
\textbf{ID} & \textbf{Dataset} & \textbf{Exercise} & \textbf{Language} & \textbf{Errors} \\
\hline \hline

\textit{Test set} \\ \hline \hline

10\_TEOS   &  FIE  &  Physics &  Java & \emph{Syntax errors} \newline Use of uninitialized variable \newline Missing variable type \\
\hline
13\_NEGF   &  FIE  &  NegaFib &  Java & \emph{Semantic errors} \newline Wrong multiplication operator \newline Incorrect parameters of recursive function call \newline \emph{other:} \newline{Forbidden library usage}\\
\hline
01\_TTBSA &  FIE  &  Triangles &  Java & \emph{Semantic errors} \newline Missing checks for illegal triangles \newline Unnecessary checks \\
\hline \hline

\multicolumn{5}{l}{\textit{Added programs for complete evaluation set}} 
 \\ \hline \hline

FIB\_01 &  Codingbat  &  Fibonacci &  Java & \emph{Semantic errors} \newline Wrong recursive call \\
\hline
FIB\_02 &  Codingbat  &  Fibonacci &  Java & \emph{No Errors} \\
\hline
PASS &  FITech  &  Password &  Dart & \emph{Semantic errors}  \newline Overwriting variable \newline Incorrect logic (missing loop) \newline Wrong datatype \\
MAX3\_01 &  TaskTracker  &  MaxOf3 &  Python & \emph{Semantic error} \newline Incorrect conditional \\
MAX3\_02 &  TaskTracker  &  MaxOf3 &  Kotlin & \emph{No Errors} \\
\hline
MAX3\_03 &  TaskTracker  &  MaxOf3 &  C++ & \emph{Style issues} \newline Unusual var names (due to post-processing) \newline Unnecessary import \\
\hline
MAX3\_04 &  TaskTracker  & MaxOf3 & Java & \emph{No Errors}  \\
\hline
MAX3\_05 &  TaskTracker  & MaxOf3  & Java & \emph{Semantic errors} \newline Missing conditionals \newline Unusual approach using subtraction in conditions \\
\hline
\end{tabular}
\end{table}

\subsubsection{Dataset Used for Evaluation}
\label{sec:ext}

To test the performance of the model with our prompt on a more diverse set of programming submissions, we expanded our dataset based on the following requirements for the exercises and corresponding solutions: \vspace*{-0.2cm}
\begin{itemize}
    \item Exercises should cover various types of tasks.
    \item Student solutions should be in different programming languages.
    \item Student solutions should have diverse types of errors.
    \item Student solutions should have different numbers of errors, ranging from zero to numerous.
    \item Student solutions should be reasonably complete. 
    \item Some of the solutions should be correct to check if the generated feedback is also valuable for correct submissions.
\end{itemize}
\vspace*{-0.2cm}

We selected a total of 11 solutions to 6 different programming exercises from existing datasets (see Table \ref{tab:tasks}) that had been used before in research~\cite{fitechdata,kiesler2022data,kiesler2022methodreport}. A new exercise, MaxOf3, was taken from the TaskTracker dataset~\cite{Lyulina2021tasktracker}. This dataset contains snapshots from 148 students solving 6 introductory programming tasks in 4 different programming languages. The data was collected within a plugin for IntelliJ-based IDEs. We selected the MaxOf3 exercise because there were both correct and incorrect submissions for it in various programming languages. The exercise was accompanied by a set of test cases (input/output) and a task description. As there was no model solution provided, we created one for each programming language.

\subsection{Prompt Design Process}

Next, we outline the iterative process of evaluating the outcomes of each prompt and how we enhanced the instructions provided within each prompt. Five iterations were needed before arriving at a prompt triggering an adequate output as a basis for an in-depth analysis. 
We defined the following conditions as a starting point for the creation of the prompt. Some of these are based on \mbox{\citeauthor{denny2021designing}}'s \mbox{\cite{denny2021designing}} guidelines for improving the readability of programming error messages.

\begin{itemize}
    \item The feedback should be concise and not too lengthy. This condition relates to the `economy of words' guideline, which means that \say{only as many words are used as are necessary to communicate the point} \cite{denny2021designing}.
    \item The feedback should be targeted at novice programmers. We assume we have no further information on the learner, such as skill level, problem-solving history, or other context. This condition relates to the `remove jargon' and `use simple vocabulary' guidelines \cite{denny2021designing}.
    \item Although feedback can be defined on many levels (high-level vs bottom-up hints), the feedback should aim at giving informative information and/or hints, but not give away the solution.
    \item The generated feedback should be only one of the specific types described in Section \ref{sec:automatedfb}.
    \item We have access to the student's code, the task description, and if applicable/available template (starter) code and test cases.
\end{itemize}

\vspace*{5cm}

\newpage

\citeauthor{denny2021designing}'s \cite{denny2021designing} fourth and final guideline `write messages in complete sentences' also applies, as LLMs typically output complete sentences.

\noindent To refine the prompt, we followed the steps below:
\vspace*{-0.3cm}
\begin{itemize}
    \item Using the existing version of the prompt, we generated feedback for the three student submissions within the test set across all six types of feedback, resulting in a total of 18 feedback messages for each iteration.
    \item Two authors analyzed the feedback messages, by indicating which feedback types were present in the output and determining the characteristics. The results were discussed by the three authors of this paper. Disagreements were discussed and resolved using a consensual approach~\cite{hill97guide,hill2005consensual}.
    \item To determine whether the prompt should be further refined, we checked whether the expected feedback type was present, if there was misleading information, or uncertainty. We then discussed improvements to the prompt and make adjustments accordingly.
\end{itemize}
\noindent
We selected OpenAI's GPT-4 model~\cite{openaiGPT4TechnicalReport2023} to generate the outputs, as one of the state-of-the-art LLMs at the time the study was conducted (March 2024). Even though its price might be a barrier for educational institutions and students, we expect this model soon to be standard (and potentially free of charge). The ChatGPT interface was utilized to generate the feedback, while a separate context window was used for each request. 
In the following paragraphs, we summarize the five iterations of the prompt design process by outlining crucial changes.

\subsubsection{First Iteration} 
We designed a first prompt based on our initial conditions, using best practices described in documents and literature \cite{10.1145/3544548.3581388, openaipromptengineering}. We included the following elements: 
\vspace*{-0.3cm}
\begin{itemize}
    \item A system prompt with instructions about the target audience, the elements given below, and an explanation of the different feedback types.
    \item The task description, translated from German into English. 
    \item The class skeleton given to the student.
    \item The student's submitted solution.
    \item A model solution.
    \item The type of feedback we want the model to generate (KR, KP, KC, KTC, KM, or KH).
\end{itemize}

When analyzing the output, we noticed several issues: First, the output was too long for the \lb{KR} feedback (290 words on average), for which we only expected a simple correct/incorrect statement with no more than 5 words (e.g., \gpt{The solution is incorrect}). In multiple instances, the response contained further feedback types in addition to the requested feedback type. Another critical issue was that the output referred to the model solution, and kept comparing the input to the model solution. This is a common problem that has been observed in previous studies~\cite{roest2023nextstep,hellasExploringResponsesLarge2023}.

To mitigate these issues, we added instructions to have only correct/incorrect as output for \lb{KR} (see line 18 in Listings \ref{prompt}), and to not provide any additional information than the desired feedback type (line 24). We also instructed the model that the feedback should never contain information about the model solution (line 25).

\subsubsection{Second Iteration}
\vspace*{-0.05cm}
We observed several improvements within the second iteration: the output was shorter, \lb{KR} feedback only gave information about correct/incorrect, and it seemed that the feedback was more related to the requested feedback types. We did not see any mentions of the model solution anymore. In terms of issues, \lb{KTC} responses sometimes just repeated the task requirements. \lb{KH} was in some cases not precise, only saying \gpt{check if correct}. \lb{KC} gives information about concepts even if there is no issue in the code relating to that concept. \lb{KP} feedback was in some cases incorrect, vague, or only \lb{KR} feedback.

We made some substantial changes to our prompt. For \lb{KC} we instructed the model to focus only on programming concepts and explain these concepts only if the issue in the submitted solution of the student is related to that concept (see line 18 in Listings \ref{prompt}). For \lb{KP} we explained that we want the model to try to divide the task into useful subgoals and give feedback about the current state of the student submission as a percentage (see line 19). For \lb{KH} feedback, we wanted the model to not only generate an instruction for the student to check if something is correct but also provide hints and suggestions on how to proceed (see line 20). For \lb{KTC} we instructed the model to only point to information in the task description if there is an issue in the student submission related to that task constraint (see line 21).

\vspace*{-0.2cm}
\subsubsection{Third Iteration}
\vspace*{-0.05cm}
In the output of the third iteration, we noticed that \lb{KM} feedback also gave hints on how to proceed, which we do not want. Moreover, the responses had become lengthier again, using verbose language which might not be appealing to novices. There were never any (code) examples or motivational words. In addition, the given type of feedback resembled our instruction. \lb{KP} gave a lengthier description instead of just a percentage, which we decided to keep so it is clear what it is based on.

In our improvements, we simplified the definition supplied with \lb{KR} about when a solution is considered correct to be either ``(1) programming solution passes all tests and contains no mistakes (2) programming solution is semantically equivalent to the model solution.'' We also told the model that the feedback has to be brief and precise. For \lb{KM} feedback, we wanted to give the student information about the error and not show hints on how to fix it. We also tried to experiment with adding illustrating examples to better support students' understanding, which we requested for \lb{KM}, \lb{KTC}, \lb{KC}, and \lb{KH}.

\vspace*{-0.2cm}
\subsubsection{Fourth Iteration}
\vspace*{-0.05cm}
The results of our previous changes did not seem to significantly improve the output. Firstly, we noticed lengthy responses. The model clearly tried to generate examples, however, we observed that the term `example' can be used in many different ways. To illustrate, the model just repeated some code from the student with issues, to be used as an `example' for a mistake. We noticed a few interesting and useful examples, such as \gpt{For example, if `a = 1`, `b = 2`, and `c = 3`, your current implementation would classify this as a SCALENE triangle, which is incorrect because these sides do not meet the triangle inequality theorem and thus cannot form a triangle}. We also noticed more code examples, but they were not always compilable. This happened when the model suggested replacing the constants defined by the student with constants from a given class:

\begin{lstlisting}[frame=none, numbers=none]
For example, in your computeP method, you directly defined BOLTZMANN and AVOGADRO 
constants:
double BOLTZMANN = 1.380649E-23;
double AVOGADRO = 6.02214076E23;    

Instead, you should utilize the constants from the \texttt{PhysicsConstants} class 
like so:
PhysicsConstants.BOLTZMANN
PhysicsConstants.AVOGADRO
\end{lstlisting}

We reworded the instruction to only show the errors for \lb{KM} and not how to fix it, and we removed the instruction to give examples. At this point, we found we could not control the example output enough, and realized that examples require more extensive attention at some point.

\vspace*{-0.2cm}
\subsubsection{Fifth iteration and final prompt}
\vspace*{-0.05cm}
In the fifth iteration, we successfully addressed the issues encountered in the fourth iteration. Yet, we observed no notable enhancements compared to the third iteration. Consequently, we chose to use this version of the prompt to assess the expanded dataset. Our final prompt is shown in Listing \ref{prompt}. \vspace*{-0.15cm}

\begin{lstlisting}[breaklines=true,frame=single,caption=Final prompt used to generate different types of feedback.,label=prompt, numbers=left]
Your task is to give feedback to a novice programmer who is working on a specific programming exercise in Java. You will be given (A) the complete task description as a LaTex Source Code, (B) the class body that is provided to the student, (C) the submitted solution of the student, (D) a model solution and (E) a desired feedback type with regard to the feedback typology described below.
Each feedback type is described with the name of the feedback type in wrapped in "" followed by a description of the feedback type wrapped in ~~

"Knowledge about task constraints (KTC)" ~ KTC are elaborated components that provide information of task rules, task constraints, and task requirements ~

"Knowledge about concepts (KC)" ~ KC are elaborated components that provide information on conceptual knowledge relevant for task processing ~

"Knowledge about mistakes (KM)" ~ KM are elaborated components that provide information on errors or mistakes ~

"Knowledge on how to proceed (KH)" ~ KH are elaborated components that provide information on procedural knowledge relevant for task processing ~

"Knowledge about performance (KP)" ~ KP provides summative feedback on the achieved performance level after doing multiple tasks or subtasks, such as '15 of 20 correct' and '85% correct' ~

"Knowledge of result/response (KR)" ~ KR is feedback that communicates whether a solution is correct or incorrect e.g. (1) programming solution passes all tests and contains no mistakes (2) programming solution is semantically equivalent to the model solution  ~

Try to make sure that the feedback you generate satisfies the following criteria:
- for feedback type KR the feedback should only contain the information if the student submission is correct or incorrect. Do not explain concepts or give information about errors or how to proceed.
- for feedback type KC only focus on programming concepts and explain these concepts only if the issue in the submitted solution of the student is related to that concept.
- for feedback type KP try to divide the task into usefull subgoals and give feedback about the current state of the student submission in percentage.
- for feedback type KH not only tell the student to check if something is correct but provide hints and suggestions on how to proceed.
- for feedback type KTC only point to information in the task description if there is an issue in the student submission related to that task constraint.
- for KM Feedback just give the student information about the error and do not describe how to fix the error.
- make sure that the generated feedback corresponds exactly to the description of the desired feedback type. Do not provide any additional information that does not match the feedback type.
- the feedback should never contain information about the model solution. The student does not know the model solution.
- your feedback has to be brief and precise.

(A) complete task description:

## INSERT Add task description here ##

(B) class body in file "##INPUT## Name of File" provided to the student:

## INSERT Add Class Body, if provided ##

(C) student's submitted solution:

## INSERT Add Student Solution ##

(D) model solution:

## INSERT Model solution ##

(E) desired feedback type:

## INSERT desired feedback type ##

\end{lstlisting} \vspace*{-0.1cm}

\subsection{Feedback Analysis}
\vspace*{-0.05cm}
To answer RQ1 and RQ2, we analyzed the generated feedback with regard to their characteristics and the actual feedback type(s), as described in Section \ref{sec:automatedfb}. For the characteristics (see Table \ref{tab:legend}), we used the deductive categories from previous work \cite{kiesler2023exploring}, extended by the two categories PERS \cite{azaiz2023aienhanced,azaiz2024feedback,roest2023nextstep} and COMPL taken from other related work~\cite{azaiz2023aienhanced,azaiz2024feedback}. The \lbb{PERS} characteristic indicates that the feedback is personalized, meaning it explicitly refers to the student's solution. The \lbb{COMPL} characteristic addresses whether the feedback complies with the task description.

\noindent The analysis was carried out in three iterations with two steps each: (1) two experts (independently) coded the generated feedback messages with the appropriate feedback types and categories, and (2) all conflicts were discussed and resolved using the consensual approach~\cite{hill97guide,hill2005consensual}. In order not to be biased when coding the generated feedback, we concealed the information on which feedback type was requested and by randomly ordering the messages. Finally, the coded data was qualitatively analyzed by identifying representative themes in the feedback, and illustrative examples.

\section{Results and Discussion}
\label{sec:res}

\subsection{RQ1: Generating Specific Feedback Types}

The results of the analysis of the generated feedback types are summarized in Table \ref{tab:coding1}. It shows that in almost all cases (63 out of 66), the desired feedback type (DFT) matches the actual feedback type (AFT) generated by the LLM. However, in 19 cases, other feedback types were additionally included in the generated feedback. It was particularly noticeable that KM feedback was often included when requesting KH feedback. However, this may be due to the fact that it is not always possible to provide information on how to proceed without (directly or indirectly) pointing to the problem in the code. The same effect was observed when requesting KTC feedback. In some cases, KTC feedback was also combined with KM or KH feedback, as it is challenging to identify problems regarding the task constraints without giving any indication of the error in the code. Regarding RQ1, we conclude that the performance of the GPT-4 model to generate specific, elaborated types of feedback is very promising.

\begin{table*}[htbp]

\centering
\caption{How ChatGPT responds to student input (RQ1). For legend see Table~\ref{tab:legend}. DFT is Desired feedback type, AFT is actual feedback type.}
\label{tab:coding1}
\small
\setlength{\tabcolsep}{0.2em}
\renewcommand{\arraystretch}{0.7}

\begin{tabular}{
  l | *{1}{c} | *{1}{c} | *{1}{c} | *{1}{c} | *{1}{c} | *{1}{c} | *{1}{c} | *{1}{c} | *{1}{c} | *{1}{c} | *{1}{c} | *{1}{c} | *{1}{c} | *{1}{c} | *{1}{c}  
}
\hline
\multicolumn{1}{ l |}{} &
\multicolumn{2}{c|}{\scriptsize FB TYPES} &
\multicolumn{6}{c|}{\scriptsize CONTENT} &
\multicolumn{3}{c|}{\scriptsize QUALITY} &
\multicolumn{4}{c}{\scriptsize OTHER}  \\
\hline 

\multicolumn{1}{ l |}{\textbf{Stud\_task}} &
\multicolumn{1}{c|}{\textbf{DFT}} &
\multicolumn{1}{c|}{\textbf{AFT}} &
\multicolumn{1}{c|}{\textbf{INFO}} &
\multicolumn{1}{c|}{\textbf{STYLE}} & 
\multicolumn{1}{c|}{\textbf{CAUSE}}& 
\multicolumn{1}{c|}{\textbf{FIX}} &
\multicolumn{1}{c|}{\textbf{CODE}} &
\multicolumn{1}{c|}{\textbf{EXA}} &

\multicolumn{1}{c|}{\textbf{COMP}} &
\multicolumn{1}{c|}{\textbf{MIS}} &
\multicolumn{1}{c|}{\textbf{UNC}} &

\multicolumn{1}{c|}{\textbf{META}} &
\multicolumn{1}{c}{\textbf{MOT}} &

\multicolumn{1}{c|}{\textbf{PERS}} &
\multicolumn{1}{c}{\textbf{COMPL}} 
\\

\hline 

\student{10\_TEOS} 
& KR           
& KR           
& \noSym           
& \noSym           
& \noSym         
& \noSym          
& \noSym       
& \noSym           
& ---            
& \noSym         
& \noSym           
& \noSym           
& \noSym           
& \yesSym           
& \noSym           

\\

\student{} 
& KTC           
& KTC           
& \noSym           
& \noSym           
& \yesSym         
& \yesSym          
& \noSym       
& \noSym           
& ---             
& \noSym         
& \noSym           
& \noSym           
& \noSym           
& \yesSym           
& \yesSym           
\\

\student{} 
& KC           
& KC KM           
& \noSym           
& \noSym           
& \yesSym         
& \yesSym          
& \noSym       
& \noSym           
& ---            
& \noSym         
& \noSym           
& \noSym           
& \noSym           
& \yesSym           
& \yesSym           
\\

\student{} 
& KM           
& KM                %
& \noSym           
& \noSym           
& \yesSym         
& \yesSym          
& \noSym       
& \noSym           
& ---            
& \noSym         
& \noSym           
& \noSym           
& \noSym           
& \yesSym           
& \yesSym           
\\

\student{} 
& KH           
& KH           
& \noSym           
& \noSym           
& \yesSym         
& \yesSym          
& \noSym       
& \noSym           
& ---            
& \noSym         
& \noSym           
& \noSym           
& \noSym           
& \yesSym           
& \yesSym           
\\
\student{} 
& KP           
& KP           
& \noSym           
& \noSym           
& \yesSym         
& \noSym          
& \noSym       
& \noSym           
& ---            
& \noSym         
& \noSym           
& \noSym           
& \noSym           
& \yesSym           
& \yesSym           
\\
\hline

\hline 

\student{13\_NEGF} 
& KR           
& KR           
& \noSym           
& \noSym           
& \noSym         
& \noSym          
& \noSym       
& \noSym           
& ---            
& \noSym         
& \noSym           
& \noSym           
& \noSym           
& \yesSym           
& \noSym           

\\

\student{} 
& KTC           
& KTC KM           
& \noSym           
& \noSym           
& \yesSym         
& \yesSym          
& \noSym       
& \noSym           
& ---             
& \noSym         
& \noSym           
& \noSym           
& \noSym           
& \yesSym           
& \yesSym           
\\

\student{} 
& KC           
& KC KM           
& \noSym           
& \noSym           
& \yesSym         
& \yesSym          
& \partSym       
& \noSym           
& ---            
& \noSym         
& \noSym           
& \noSym           
& \noSym           
& \yesSym           
& \yesSym           
\\

\student{} 
& KM           
& KM                %
& \noSym           
& \noSym           
& \yesSym         
& \yesSym          
& \partSym       
& \noSym           
& ---            
& \noSym         
& \noSym           
& \noSym           
& \noSym           
& \yesSym           
& \yesSym           
\\

\student{} 
& KH           
& KH KM          
& \noSym           
& \noSym           
& \yesSym         
& \yesSym          
& \partSym       
& \noSym           
& ---            
& \noSym         
& \noSym           
& \noSym           
& \noSym           
& \yesSym           
& \yesSym           
\\

\student{} 
& KP           
& KP KM          
& \noSym           
& \noSym           
& \yesSym         
& \yesSym          
& \noSym       
& \noSym           
& ---            
& \noSym         
& \noSym           
& \noSym           
& \noSym           
& \yesSym           
& \yesSym           
\\
\hline

\hline 

\student{01\_TTBSA} 
& KR           
& KR           
& \noSym           
& \noSym           
& \noSym         
& \noSym          
& \noSym       
& \noSym           
& ---            
& \noSym         
& \noSym           
& \noSym           
& \noSym           
& \yesSym           
& \noSym           
\\

\student{} 
& KTC           
& KTC           
& \noSym           
& \noSym           
& \yesSym         
& \yesSym          
& \noSym       
& \noSym           
& ---             
& \noSym         
& \noSym           
& \noSym           
& \noSym           
& \yesSym           
& \yesSym           
\\

\student{} 
& KC           
& KC KM           
& \noSym           
& \noSym           
& \yesSym         
& \yesSym          
& \noSym       
& \noSym           
& ---            
& \noSym         
& \noSym           
& \noSym           
& \noSym           
& \yesSym           
& \yesSym           
\\

\student{} 
& KM           
& KM                %
& \noSym           
& \noSym           
& \yesSym         
& \yesSym          
& \noSym       
& \noSym           
& ---            
& \noSym         
& \noSym           
& \noSym           
& \noSym           
& \yesSym           
& \yesSym           
\\

\student{} 
& KH           
& KH          
& \noSym           
& \noSym           
& \yesSym         
& \yesSym          
& \noSym       
& \noSym           
& ---            
& \noSym         
& \noSym           
& \noSym           
& \noSym           
& \yesSym           
& \yesSym           
\\
\student{} 
& KP           
& KP KM          
& \noSym           
& \noSym           
& \yesSym         
& \noSym          
& \noSym       
& \noSym           
& ---            
& \noSym         
& \noSym           
& \noSym           
& \noSym           
& \yesSym           
& \yesSym           
\\
\hline

\hline 

\student{FIB\_01} 
& KR           
& KR           
& \noSym           
& \noSym           
& \noSym         
& \noSym          
& \noSym       
& \noSym           
& ---            
& \noSym         
& \noSym           
& \noSym           
& \noSym           
& \yesSym           
& \noSym           

\\

\student{} 
& KTC           
& KTC           
& \noSym           
& \noSym           
& \yesSym         
& \noSym          
& \partSym       
& \noSym           
& ---             
& \noSym         
& \noSym           
& \noSym           
& \noSym           
& \yesSym           
& \yesSym           
\\

\student{} 
& KC           
& KC KM           
& \noSym           
& \noSym           
& \yesSym         
& \noSym          
& \noSym       
& \noSym           
& ---            
& \noSym         
& \noSym           
& \noSym           
& \noSym           
& \yesSym           
& \yesSym           
\\

\student{} 
& KM           
& KM KH           %
& \noSym           
& \noSym           
& \yesSym         
& \yesSym          
& \partSym       
& \noSym           
& ---            
& \noSym         
& \noSym           
& \noSym           
& \noSym           
& \yesSym           
& \yesSym           
\\

\student{} 
& KH           
& KH KM           
& \noSym           
& \noSym           
& \yesSym         
& \yesSym          
& \partSym       
& \noSym           
& ---            
& \noSym         
& \noSym           
& \noSym           
& \noSym           
& \yesSym           
& \yesSym           
\\
\student{} 
& KP           
& KP KM           
& \noSym           
& \noSym           
& \yesSym         
& \noSym          
& \partSym       
& \noSym           
& ---            
& \noSym         
& \noSym           
& \noSym           
& \yesSym           
& \yesSym           
& \yesSym           
\\
\hline

\student{FIB\_02 \checkmark} 
& KR           
& KR           
& \noSym           
& \noSym           
& \noSym         
& \noSym          
& \noSym       
& \noSym           
& ---            
& \noSym         
& \noSym           
& \noSym           
& \noSym           
& \yesSym           
& \noSym           

\\

\student{} 
& KTC           
& KTC           
& \noSym           
& \noSym           
& \noSym         
& \noSym          
& \noSym       
& \noSym           
& ---             
& \noSym         
& \noSym           
& \noSym           
& \noSym           
& \noSym           
& \yesSym           

\\

\student{} 
& KC           
& KC           
& \noSym           
& \noSym           
& \noSym         
& \noSym          
& \noSym       
& \noSym           
& ---            
& \noSym         
& \noSym           
& \noSym           
& \noSym           
& \yesSym           
& \yesSym           

\\

\student{} 
& KM           
& KM KH           %
& \noSym           
& \noSym           
& \yesSym         
& \yesSym          
& \partSym       
& \noSym           
& ---            
& \yesSym         
& \noSym           
& \noSym           
& \noSym           
& \yesSym           
& \yesSym           

\\

\student{} 
& KH           
& KH KC           
& \noSym           
& \noSym           
& \noSym         
& \noSym          
& \noSym       
& \noSym           
& ---            
& \noSym         
& \noSym           
& \noSym           
& \noSym           
& \noSym           
& \noSym           

\\

\student{} 
& KP           
& KP KM           
& \noSym           
& \noSym           
& \noSym         
& \noSym          
& \noSym       
& \noSym           
& ---            
& \noSym         
& \noSym           
& \noSym           
& \noSym           
& \yesSym           
& \yesSym           
\\
\hline

\student{PASS\_01} 
& KR           
& KR           
& \noSym           
& \noSym           
& \noSym         
& \noSym          
& \noSym       
& \noSym           
& ---            
& \noSym         
& \noSym           
& \noSym           
& \noSym           
& \yesSym           
& \noSym           

\\

\student{} 
& KTC           
& KTC KM KH          
& \noSym           
& \noSym           
& \yesSym         
& \yesSym          
& \noSym       
& \noSym           
& ---             
& \noSym         
& \noSym           
& \noSym           
& \noSym           
& \yesSym           
& \yesSym           

\\

\student{} 
& KC           
& KC, KM           
& \noSym           
& \noSym           
& \yesSym         
& \noSym          
& \partSym       
& \noSym           
& ---            
& \noSym         
& \noSym           
& \noSym           
& \noSym           
& \yesSym           
& \yesSym           

\\

\student{} 
& KM           
& KM           %
& \noSym           
& \noSym           
& \yesSym         
& \noSym          
& \partSym       
& \noSym           
& ---            
& \noSym         
& \noSym           
& \noSym           
& \noSym           
& \yesSym           
& \yesSym           

\\

\student{} 
& KH           
& KH           
& \noSym           
& \noSym           
& \noSym         
& \yesSym          
& \noSym       
& \noSym           
& ---            
& \noSym         
& \noSym           
& \noSym           
& \noSym           
& \yesSym           
& \yesSym           

\\

\student{} 
& KP           
& KP KM           
& \noSym           
& \noSym           
& \yesSym         
& \noSym          
& \noSym       
& \noSym           
& ---            
& \noSym         
& \noSym           
& \noSym           
& \noSym           
& \yesSym           
& \yesSym           
\\
\hline

\student{MAX3\_01} 
& KR           
& KR           
& \noSym           
& \noSym           
& \noSym         
& \noSym          
& \noSym       
& \noSym           
& ---            
& \yesSym         
& \noSym           
& \noSym           
& \noSym           
& \yesSym           
& \noSym           

\\

\student{} 
& KTC           
& KTC          
& \noSym           
& \noSym           
& \noSym         
& \noSym          
& \noSym       
& \noSym           
& ---             
& \yesSym         
& \noSym           
& \noSym           
& \noSym           
& \yesSym           
& \yesSym           

\\

\student{} 
& KC           
& KC KM           
& \noSym           
& \noSym           
& \yesSym         
& \yesSym          
& \noSym       
& \noSym           
& ---            
& \noSym         
& \noSym           
& \noSym           
& \noSym           
& \yesSym           
& \yesSym           

\\

\student{} 
& KM           
& KM           %
& \noSym           
& \noSym           
& \yesSym         
& \noSym          
& \partSym       
& \yesSym           
& ---            
& \yesSym         
& \noSym           
& \noSym           
& \noSym           
& \yesSym           
& \yesSym           

\\

\student{} 
& KH           
& KH           
& \noSym           
& \yesSym           
& \noSym         
& \yesSym          
& \noSym       
& \noSym           
& ---            
& \noSym         
& \noSym           
& \noSym           
& \noSym           
& \yesSym           
& \yesSym           

\\

\student{} 
& KP           
& KP KM           
& \noSym           
& \noSym           
& \noSym         
& \yesSym          
& \noSym       
& \noSym           
& ---            
& \yesSym         
& \noSym           
& \noSym           
& \noSym           
& \yesSym           
& \yesSym           
\\
\hline

\student{MAX3\_02 \checkmark} 
& KR           
& KR           
& \noSym           
& \noSym           
& \noSym         
& \noSym          
& \noSym       
& \noSym           
& ---            
& \noSym         
& \noSym           
& \noSym           
& \noSym           
& \yesSym           
& \noSym           

\\

\student{} 
& KTC           
& KTC          
& \noSym           
& \noSym           
& \yesSym         
& \yesSym          
& \noSym       
& \noSym           
& ---             
& \yesSym         
& \noSym           
& \noSym           
& \noSym           
& \yesSym           
& \noSym           

\\

\student{} 
& KC           
& KC           
& \noSym           
& \yesSym           
& \noSym         
& \noSym          
& \partSym       
& \noSym           
& ---            
& \noSym         
& \noSym           
& \noSym           
& \noSym           
& \yesSym           
& \yesSym           

\\

\student{} 
& KM           
& KM KH           %
& \noSym           
& \yesSym           
& \yesSym         
& \yesSym          
& \partSym       
& \noSym           
& ---            
& \yesSym         
& \noSym           
& \noSym           
& \noSym           
& \yesSym           
& \yesSym           

\\

\student{} 
& KH           
& KH           
& \noSym           
& \yesSym           
& \noSym         
& \yesSym          
& \noSym       
& \noSym           
& ---            
& \yesSym         
& \noSym           
& \noSym           
& \noSym           
& \yesSym           
& \yesSym           

\\

\student{} 
& KP           
& KP KM           
& \noSym           
& \noSym           
& \noSym         
& \noSym          
& \noSym       
& \noSym           
& ---            
& \yesSym         
& \noSym           
& \noSym           
& \noSym           
& \yesSym           
& \yesSym           
\\
\hline

\student{MAX3\_03 \checkmark} 
& KR           
& KR           
& \noSym           
& \noSym           
& \noSym         
& \noSym          
& \noSym       
& \noSym           
& ---            
& \noSym         
& \noSym           
& \noSym           
& \noSym           
& \yesSym           
& \noSym           

\\

\student{} 
& KTC           
& KTC          
& \noSym           
& \noSym           
& \noSym         
& \noSym          
& \noSym       
& \noSym           
& ---             
& \yesSym         
& \noSym           
& \noSym           
& \noSym           
& \yesSym           
& \yesSym           

\\

\student{} 
& KC           
& KC           
& \noSym           
& \noSym           
& \noSym         
& \noSym          
& \noSym       
& \noSym           
& ---            
& \noSym         
& \noSym           
& \noSym           
& \noSym           
& \yesSym           
& \yesSym           

\\

\student{} 
& KM           
& KM KH           %
& \noSym           
& \yesSym           
& \yesSym         
& \yesSym          
& \noSym       
& \noSym           
& ---            
& \noSym         
& \noSym           
& \noSym           
& \noSym           
& \yesSym           
& \yesSym           

\\

\student{} 
& KH           
& KH           
& \noSym           
& \yesSym           
& \yesSym         
& \yesSym          
& \noSym       
& \noSym           
& ---            
& \yesSym         
& \noSym           
& \noSym           
& \noSym           
& \yesSym           
& \yesSym           

\\

\student{} 
& KP           
& KP KM           
& \noSym           
& \noSym           
& \noSym         
& \noSym          
& \noSym       
& \noSym           
& ---            
& \yesSym         
& \noSym           
& \noSym           
& \noSym           
& \yesSym           
& \yesSym           
\\
\hline

\student{MAX3\_04 \checkmark} 
& KR           
& KR           
& \noSym           
& \noSym           
& \noSym         
& \noSym          
& \noSym       
& \noSym           
& ---            
& \noSym         
& \noSym           
& \noSym           
& \noSym           
& \yesSym           
& \noSym           

\\

\student{} 
& KTC           
& KTC          
& \noSym           
& \noSym           
& \yesSym         
& \noSym          
& \noSym       
& \noSym           
& ---             
& \yesSym         
& \noSym           
& \noSym           
& \noSym           
& \yesSym           
& \yesSym           

\\

\student{} 
& KC           
& KC           
& \noSym           
& \noSym           
& \noSym         
& \noSym          
& \noSym       
& \noSym           
& ---            
& \noSym         
& \noSym           
& \noSym           
& \noSym           
& \yesSym           
& \yesSym           

\\

\student{} 
& KM           
& KM           %
& \noSym           
& \yesSym           
& \yesSym         
& \noSym          
& \noSym       
& \noSym           
& ---            
& ?         
& \noSym           
& \noSym           
& \noSym           
& \yesSym           
& \yesSym           

\\

\student{} 
& KH           
& KH           
& \noSym           
& \yesSym           
& \noSym         
& \yesSym          
& \noSym       
& \noSym           
& ---            
& \noSym         
& \noSym           
& \noSym           
& \noSym           
& \yesSym           
& \yesSym           

\\

\student{} 
& KP           
& KP           
& \noSym           
& \noSym           
& \noSym         
& \noSym          
& \noSym       
& \noSym           
& ---            
& \noSym         
& \noSym           
& \noSym           
& \noSym           
& \yesSym           
& \yesSym           
\\
\hline

\student{MAX3\_05} 
& KR           
& KR           
& \noSym           
& \noSym           
& \noSym         
& \noSym          
& \noSym       
& \noSym           
& ---            
& \noSym         
& \noSym           
& \noSym           
& \noSym           
& \yesSym           
& \noSym           

\\

\student{} 
& KTC           
& KM KH          
& \noSym           
& \noSym           
& \yesSym         
& \yesSym          
& \noSym       
& \noSym           
& ---             
& \noSym         
& \noSym           
& \noSym           
& \noSym           
& \yesSym           
& \yesSym           

\\

\student{} 
& KC           
& KC           
& \noSym           
& \yesSym           
& \noSym         
& \noSym          
& \partSym       
& \noSym           
& ---            
& \noSym         
& \noSym           
& \noSym           
& \noSym           
& \yesSym           
& \yesSym           

\\

\student{} 
& KM           
& KM           %
& \noSym           
& \noSym           
& \yesSym         
& \noSym          
& \noSym       
& \noSym           
& ---            
& \noSym         
& \noSym           
& \noSym           
& \noSym           
& \yesSym           
& \yesSym           

\\

\student{} 
& KH           
& KH KM           
& \noSym           
& \noSym           
& \yesSym         
& \yesSym          
& \noSym       
& \noSym           
& ---            
& \noSym         
& \noSym           
& \yesSym           
& \noSym           
& \yesSym           
& \yesSym           

\\

\student{} 
& KP           
& KP KM           
& \noSym           
& \noSym           
& \noSym         
& \noSym          
& \noSym       
& \noSym           
& ---            
& \noSym         
& \noSym           
& \noSym           
& \noSym           
& \yesSym           
& \yesSym           
\\
\hline

\multicolumn{13}{c}{\yesSym: Yes \quad \noSym: No \quad \partSym: Snippet \quad ---: Not applicable \quad \checkmark: Correct Solution}

\end{tabular}

\end{table*}

\subsection{RQ2: Characteristics of Generated Feedback}

In this section, we describe the characteristics we found in the generated feedback messages and conduct a qualitative analysis in which we highlight certain themes identified in the data.

\subsubsection{Dealing with correct programs}

Several correct student programs were used as input to the LLM to find out the extent to which GPT-4 can recognize these and provide adequate feedback. For example, consider the student program in Listing \ref{lst:kotlin1}, which is a correct \texttt{Kotlin} solution to the MaxOf3 exercise. The test cases for this exercise were shown in the form of text as `$1\; 2\; 3 \rightarrow 3$'. However, the submissions required the input to be inserted line-by-line. The \lb{KTC} feedback confuses these by stating \gpt{... your current implementation reads three integers in a single line, which may not align with the test cases provided in the task description where inputs are expected line by line}, with the advice to \gpt{review the task description and adjust your input reading mechanism to match the expected format of the test cases.} Therefore, we consider this feedback to be misleading. We observed similar misleading feedback for a correct \texttt{C++} and a correct \texttt{Java} submission to the same task. For the \texttt{C++} program, the LLM responded: \gpt{Your program should output only the largest number, without any preceding text}. However, the student had already done this and only printed the largest number (which was also the case in the \texttt{Java} submission).

\begin{lstlisting}[language=Java,caption=Correct Kotlin submission to MaxOf3.,label=lst:kotlin1]
import java.util.*
import kotlin.math.max

fun main() {
    val main_v1 = Scanner(System.`in`)
    val main_v2 = main_v1.nextInt()
    val main_v3 = main_v1.nextInt()
    val main_v4 = main_v1.nextInt()
    println(max(max(main_v2, main_v3), main_v4))
}  
\end{lstlisting}

The other feedback types for the \texttt{Kotlin} submission (Listing \ref{lst:kotlin1}) all contain advice on \lbb{style}. It is mentioned that the code could be made \gpt{cleaner} and more idiomatic to \texttt{Kotlin} by using \snpt{readLine()}. This is perfectly fine advice; however, in the \lb{KM} feedback the term `mistake' is used for this issue, which is misleading since it is not really problematic. This is also the case for the \lb{KP} feedback, which awards a score of 85\% for this correct program that would not need changes.

The \lb{KC} feedback for a correct \texttt{C++} submission gave confusing feedback on the student using the wrong header file for the built-in max-function, while this import was not even necessary. The \lb{KM} feedback for the same solution acknowledges this issue as well, but phrases it as a style suggestion: \gpt{Removing unused libraries helps to keep the codebase more maintainable and potentially reduces the compilation time.}

For a correct but somewhat overly complex solution to MaxOf3 in \texttt{Java}, GPT-4 suggested useful improvements: \gpt{consider structuring your conditional statements to more clearly identify the largest number among the three inputs. Start by comparing the first number with the second and third numbers to determine if it's the largest. If it's not, move on to compare the second number with the third. This way, you can avoid redundant comparisons and make your logic more straightforward.}

\subsubsection{Misleading feedback}

One of the major issues pointed out by previous studies is the regular occurrence of misleading advice. We still noticed such issues in the feedback, but misleading information seemed to occur less frequently (compared to~\cite{kiesler2023exploring}). One example of misleading feedback pointed out \gpt{a minor inconsistency in the use of braces in the `if-else` blocks, where one of the `else` statements does not use braces for a single statement, unlike the others}. This was not the case in the actual student code. Therefore, it may be very confusing feedback for a novice learner. In the next sections, we highlight examples of misleading feedback for specific feedback types. 

\subsubsection{Feedback on overall performance (\lb{KP})}

Although our focus was on formative feedback, we were also interested in getting feedback on performance in general. While \lb{KP} is usually considered a simple feedback type, with a number or percentage indicating progress, we decided to generate more elaborate reports on student's progress. We used the concept of `subgoals'~\cite{margulieux2016employing}, which has been studied extensively in computing education research. The results for this feedback type were quite good, although we did not always agree with the percentage score given by the model, often giving too many points for minor subgoals. 

For the MaxOf3 exercise, GPT-4 divided the problem into three reasonable subgoals in most cases, for example, \gpt{(1) reading inputs correctly, (2) comparing the numbers accurately, and (3) printing the correct output.} For most correct submissions, the expected score of 100\% was given, together with a description of how the solution meets the subgoals. However, we also spotted an instance where a score of 80\% was displayed for a correct solution.

\subsubsection{Feedback on concepts (\lb{KC})}

We instructed the model to focus on \textit{programming} concepts, as opposed to other concepts related to the task description, such as mathematical formulas. We expected these concepts to appear in feedback on (semantic) mistakes. In most cases, this worked out well. For the correct solutions, GPT-4 usually made general remarks about a concept, possibly referring to how it is applied in the student's solution. For the incorrect solutions, the concepts were related to actual mistakes and, in some cases, contained some information on how to fix them.

As an example, for an incorrect solution to the NegaFib exercise, the output for \lb{KC} focused on the concept of recursion, giving rather a long explanation, including \gpt{each function call should ideally break down the problem into smaller, more manageable subproblems that resemble the original problem}. This feedback also contained some \lbb{CAUSE} and \lbb{FIX} elements, to point out the mistake made related to recursion.

For a \texttt{Kotlin} solution to MaxOf3, GPT-4 addressed the concept of the library function \texttt{max}: \gpt{You've nested two \texttt{max} functions to compare three values: \texttt{max(max(main\_v2, main\_v3), main\_v4)}. This is a valid approach to find the largest of three numbers...}. 
In addition, it stated \gpt{However, it's important to understand that \texttt{Kotlin} also provides conditional expressions, such as \texttt{if} statements and \texttt{when} expressions, that can be used to solve this problem in a more readable or structured way, especially when dealing with more complex conditions or a larger set of numbers.} Other concepts that GPT-4 brought up were \texttt{C++} header files, comparison operators, constants, and variable scope.

Another subtype of KC deals with giving examples. None of the generated feedback messages contained complete code examples. This is hardly surprising, as we explicitly requested in our prompt that no solution should be provided. However, code snippets were generated in many explanations (especially for types KM, KH, and KTC). These consisted, for example, of references to parts of the code. As there were no complete code examples in the feedback messages, the ``COMP'' category was obsolete. It was noticeable that there were hardly any examples in the feedback messages. By examples, we mean illustrative examples that, for example, underpin the explanation of a concept or make it easier to understand. We explicitly did not code references to parts of the code as examples.

\subsubsection{Feedback on task constraints (\lb{KTC})}

This type of feedback encompasses hints on task requirements, such as enforcing the use of a particular language construct or prohibiting something. Another subcategory contains hints with general information on how to approach the exercise, not being specific to the student's current work. In some of the outputs we noticed GPT-4 giving feedback of this kind, such as for the Physics task: \gpt{The task specifically requires you to use the constants defined in the \texttt{PhysicsConstants} class for these values to ensure consistency and maintainability of the code.}, and the NegaFibonacci: \gpt{The task explicitly requires the use of the generalized Fibonacci sequence formula for negaFibonacci, without resorting to alternative formulas or approaches.}

In other cases, the feedback referred to \textit{functional} constraints that can be derived from the task description, which we assign to \lb{KM} feedback on solution errors. For the correct solutions to the MaxOf3 exercise, which did not have any particular constraints, the generated feedback mentioned something about checking the output format. It even gave misleading suggestions that \gpt{Your program should output only the largest number, without any preceding text}, while this was not the case. In several cases, we also noticed that GPT-4 simply rephrased the task description.


\subsubsection{Feedback on mistakes (\lb{KM})}

There are several subtypes to this category: feedback on compiler errors, solution errors, test failures, style issues, and performance issues. This feedback can be short, but we mostly observed elaborate error descriptions. We instructed GPT not to give information on how to fix the error, but in some cases, this was still included. In most cases, GPT pointed out actual errors but made a few mistakes. For example, for the program below in Listing \ref{lst:java2} the feedback was \gpt{Your solution correctly implements the Fibonacci sequence; however, there's a mistake in the order of your base case checks. You should check for \texttt{n == 0} before \texttt{n == 1} to adhere to the sequence's definition starting from 0.} This might indeed look somewhat nicer, but does not have any consequences for the method's correctness.

\newpage
\begin{lstlisting}[language=Java,caption=Correct Java submission to Fibonacci (FIB02).,label=lst:java2]
public int fibonacci(int n) {
    if(n==1) return 1; 
    if(n==0) return 0;
    return fibonacci(n-2) + fibonacci(n-1);
}
\end{lstlisting}

In general, we observed more often that GPT used the term `mistake' for issues related to style, which did not influence the correctness of the code. We consider this to be misleading for students, because they will think they did something wrong, instead of just giving a hint for an improvement. This is probably due to the fact that we call this type `mistake', but we actually mean it in a broader sense (e.g., to include style issues).

\lb{KM} inherently always contains a \lbb{CAUSE} explaining where or what the error is and is therefore also personalized (\lbb{PERS}). 
In some cases, style suggestions were given. It was noticeable that this was particularly prevalent in feedback messages on correct tasks. 

\vspace*{-0.2cm}
\subsubsection{Feedback on how to proceed (\lb{KH})}
\vspace*{-0.05cm}
The CAUSE and FIX categories are closely linked to the KM and KH feedback types. If KM feedback was requested, CAUSE was always included. The same applies to KH feedback. This is not surprising, as \citeauthor{kiesler2023exploring} \cite{kiesler2023exploring} had derived these two categories from KM and KH feedback. 
In many cases, the KH feedback gave valuable instructions -- formulated as individual steps -- on how to address the problems in the code. In some cases, the feedback was less valuable as it contained information that in principle corresponds to a ``think harder'' statement, e.g., simply asking the learner to ensure that a particular specification was correct without concrete help on how to achieve this (\gpt{Ensure your method correctly implements the formula provided in the task description.}).

\vspace*{-0.2cm}
\subsubsection{Dealing with logic}
\vspace*{-0.05cm}
It is well known that LLMs struggle with logical reasoning~\cite{creswell2022selectioninference}. In one \texttt{Java} solution to MaxOf3 the student had the unusual approach of comparing numbers by subtracting them and checking for a positive or negative result (see Listing \ref{lst:java1}). GPT-4 was not always precise in pointing out the problem, noting that \gpt{your solution currently lacks a branch to handle the situation where the first number is less than the second but greater than the third}, which is not a particular case that needs to be dealt with. Moreover, the model added that \gpt{adding this condition will make your solution more comprehensive}, which does not seem like a useful goal.

\begin{lstlisting}[language=Java,caption=Fragment of incorrect Java submission to MaxOf3 (omitting initialization of variables and scanner.,label=lst:java1]
//..
main_v1 = main_v7.nextInt();
main_v2 = main_v7.nextInt();
main_v3 = main_v7.nextInt();
main_v4 = main_v1 - main_v2;
if (main_v4 <=0){
    if ((main_v2 - main_v3)<=0){
        System.out.println(main_v3);
    } else {
        System.out.println(main_v2);
    }
}
\end{lstlisting}

\subsubsection{Language and tone of the feedback}

In our prompt, we explicitly instruct ChatGPT to give feedback to a \textit{novice} programmer. Additionally, our exercises are clearly aimed at beginners. We noticed that the generated feedback was suitable for the target audience. However, there were some instances of specific terminologies, such as \gpt{The submitted solution has an error related to variable shadowing and data type mismatch}, but in this case, the errors were explained afterward.

As in a previous study~\cite{kiesler2023exploring}, very few feedback messages contained motivational components. Another study \cite{roest2023nextstep} also identified few compliments in feedback messages, although observed mostly a `friendly' tone. We only coded the MOT category if there was a clear intention to motivate (e.g. ``Well done''). The feedback sometimes provided information about (partial) correct solutions (e.g., ``Your implementation of the base cases is correct''). While these cases may serve the purpose of motivating the learner, we did not label them as MOT because we could not determine the 'real intention' behind the generated messages. The only motivating text we noticed was \gpt{Your current solution has made a good attempt at implementing the Fibonacci sequence, but it's not fully aligned with the requirements of the Fibonacci calculation...}

\citeauthor{kiesler2023exploring} \cite{kiesler2023exploring} also reported feedback messages asking for more information or showing uncertainty when providing feedback to learners' submissions. In contrast, we did not observe any feedback messages requesting additional information or responses with uncertainty. This may be attributed to the more comprehensive context provided in the prompt during generation. We do not assume that this improvement is due to the fact that we used a newer model (GPT-4 instead of GPT-3.5).

Fortunately, we did not find any instances of derogatory or demeaning tone in the generated feedback messages. The language was consistently grammatically correct and expressed in a friendly or neutral tone.

\vspace*{-0.1cm}
\section{Limitations}
\label{sec:lim}
\vspace*{-0.2cm}

The results presented in this paper are based on the evaluation of outputs from a single LLM (GPT-4). The evaluation of such feedback by experts is a time-consuming task, so we had to decide to either (a) evaluate multiple models on a small set of different tasks, or (b) evaluate a more heterogeneous set of tasks and responses from a single model. We chose the second option as the GPT-4 model seemed to be the most promising and investigated model (at the time of the study). In addition, we assume it is more valuable to investigate multiple programming languages, topics, and error types, and how one system responds to them. 

Another limitation is due to the selected programming exercises as part of our dataset, which mostly concern mathematical functions. Although this is common in the first few weeks of introductory programming courses, it does not cover all possible types of tasks. Therefore, we have to be careful to generalize our results to other (non-mathematical) task domains in introductory programming.

\section{Implications and Recommendations}
\label{sec:implications}

\vspace*{-0.2cm}
\subsection{Implications for Educators}
\vspace*{-0.01cm}
Incorporating LLMs for programming in educational settings can be beneficial for improving the feedback loop, alleviating teachers from frequently having to provide individual feedback. The potential effectiveness of LLM-driven feedback, however, highly depends on the precision and contextualization of the used prompt. Previous research has highlighted the limitations of overly simplistic prompts. Simply providing student code with a general request for feedback may lead to (partially) incorrect or misleading responses, or the model requesting more information~\cite{kiesler2023exploring}. The present study reveals that it is possible to generate specific feedback types tailored to the investigated tasks.

\vspace*{-0.02cm}
Consequently, we can assume two use cases for (online) course scenarios: (1) The educator selects a suitable feedback type and submits the respective prompt(s) to the model. (2) Learners make their requests to the model to generate specific feedback types (e.g., during the programming process). In both cases, some guidance or introduction is needed on what to consider when requesting feedback from the model. Studies have shown that requests to the model without a certain level of basic understanding lead to poorer results~\cite{prather2023weird,jeuring2023what}. If students can use predefined prompt(s) (i.e., those investigated in the present paper), students can simply select the desired feedback type. For example, if they only want a hint on how to proceed to correct the faulty code (KH) or information about the error and where it is located in the code (KM). Or does the learner want a report on their progress of the task (KP)? The choice would be up to them. Instead of requiring learners to adapt the prompt themselves, the feedback request should be facilitated through a simplified method, such as a button sending the appropriate prompt to request and display the desired feedback, avoiding the need for direct interaction of the learner with the prompt (see \cite{lohrAdaptiveLearningSystems2024}). Unfortunately, the LLM still occasionally hallucinates and produces misleading information, which is a major concern -- especially if the output is merely available to the learners and not reflected or critically discussed. Learners need to become aware that the feedback they receive may sound good but actually contains misleading information. Educators should, therefore, support students in that learning process. 

\vspace*{-0.3cm}
\subsection{Implications for Tool Developers}
\vspace*{-0.05cm}

The results of this study have shown that the generation of feedback for programming tasks using LLMs has the potential to improve educational systems. The major advantage over previous approaches is that LLMs do not have to be trained by users themselves on large amounts of data, which saves considerable development efforts. Nevertheless, we believe an approach based solely on LLMs is not the most promising. Our results show that even simple feedback such as \lb{KR} is not always correct, claiming that a correct submission was incorrect, and vice versa. Hybrid solutions may be more promising, in which, for example, test cases are combined with requests to the LLM. Then test cases are evaluated and their outcomes are inserted into the prompt as information. We assume that in this case, the probability of misleading information in KR is significantly lower and the KP feedback will also be of higher quality. We have seen that in the case of (correctly diagnosed) solutions, the feedback often contained hints on how to improve the code in terms of readability and structure (style). A hybrid approach with test cases and LLMs could therefore create further opportunities to adapt the feedback even more specifically to the status of the current submission -- e.g. \gpt{Give only STYLE suggestions iff the submitted solution passes all tests}.

\vspace*{-0.3cm}
\subsection{Implications for Researchers}
\vspace*{-0.05cm}

In this work, we have created and analyzed a dataset consisting of LLM-generated feedback on authentic student submissions in an introductory programming course. Most of the feedback corresponded to the feedback types we intended to generate, but we only examined the quality of the feedback by looking at certain characteristics. In future work, researchers could investigate students' and experts' perspectives on the helpfulness of the generated feedback. So far, we generated one type of feedback for each request. In a real-world scenario, feedback from experts does not exclusively consist of only one type. It is therefore worth investigating whether and how several types of feedback can be combined and generated in conjunction. In terms of expanding the feedback variants, researchers can investigate how to generate even more specific types of feedback, such as the subtypes from the applied feedback classification~\cite{keuning2018, narciss2006}. Another pathway is to investigate different levels of detail, from high-level hints to concrete suggestions, and to personalize the feedback even more to students' preferences. The provided prompt could also be adapted for more complex tasks.

\section{Conclusion}
\label{sec:concl}

In this study, we investigated how LLMs could be instructed to generate feedback of different types (according to an existing feedback classification) for solutions to introductory programming exercises. We incrementally developed a prompt in five iterations, arriving at a version that generated the requested feedback types in most cases. We analyzed the output of the final prompt using GPT-4 for each of the six feedback types and 11 student solutions. The student submissions for a total of six programming exercises contain various kinds of errors and are written in four different programming languages (\texttt{Java}, \texttt{Python}, \texttt{C++}, and \texttt{Kotlin}). We qualitatively analyzed the resulting 66 feedback messages by categorizing the resulting feedback type(s), and recording the 13 characteristics defined in prior work, resulting in the analysis of 858 features. 

We found that our prompt generated the desired feedback type in most of the cases (63 out of 66). At the same time, we identified elements that were not requested. The prompt for \lb{KR} feedback always provided only a correct/incorrect evaluation, which in most cases matched the actual (in)correctness of the solution. The \lb{KTC} feedback prompt mostly generated relevant task constraints, such as the absence of a required concept, and to give feedback accordingly. However, we noticed that functional constraints were generated, which we would have expected as part of \lb{KM} feedback. The \lb{KC} feedback prompt introduced programming concepts identified in the exercises students struggled with. It was noted in combination with \lb{KM} elements. In general, the \lb{KM} feedback was very good at pointing out the errors in student programs, showing an improvement upon earlier studies with older models. In some cases, the fixes to these problems were also included (\lb{KH}). Finally, we observed that GPT-4 was good at generating \lb{KP} feedback by dividing the problem into subgoals. Nonetheless, the resulting percentages indicating correctness did not always match our estimations.

Earlier work has shown that existing learning environments lack diversity in the types of feedback they provide, usually resorting to types that are easy to automate, e.g., by showing test case results. With this work, we propose the prompting of LLMs in a controlled way, to give richer feedback to students not limited to scores and lists of mistakes. LLMs, such as GPT-4, can generate explanations for relevant programming concepts, point to task requirements, hints on how to address mistakes and information about students' progress throughout the task. This way, educators and learners are enabled to apply LLMs in a way that can be beneficial, providing appropriate guardrails, as opposed to giving them the freedom to request a solution to the task by asking ChatGPT and causing over-reliance on generative AI tools \cite{prather2023wgfullreport}. 
We further described implications for educators, researchers, and tool developers, advocating for a hybrid approach to combine LLM-based feedback with established methods and human instruction. In addition, this research will support researchers and teachers with the means to better study the effects of different types of feedback in the future.

\newpage
\bibliography{literature}

\end{document}